\documentclass[letterpaper, 10 pt, conference]{ieeeconf}  

\IEEEoverridecommandlockouts                              

\usepackage{multirow}
\usepackage{tikz}
\usepackage{colortbl}
\usepackage{makecell}

\def\BibTeX{{\rm B\kern-.05em{\sc i\kern-.025em b}\kern-.08em
    T\kern-.1667em\lower.7ex\hbox{E}\kern-.125emX}}

\usepackage[table]{xcolor}
\usepackage{colortbl}
\usepackage{graphicx}
\usepackage{caption}
\usepackage{amsmath} 
\usepackage{hyperref}
\usepackage{cleveref}
\usepackage[mathscr]{euscript}
\usepackage[table]{xcolor}
\usepackage{booktabs}

\title{\LARGE \bf
RoDyn: Taming Interactive Robot-Dynamic 2.5D World Model \\ for Robotic Manipulation
}

\author{Chuanrui Zhang$^{1}$, Zhengxian Wu$^{2}$, Guanxing Lu$^{2}$, Yansong Tang$^{2}$, Ziwei Wang$^{1}$ \\
$^{1}$Nanyang Technological University, Singapore \qquad $^{2}$Tsinghua University, Beijing, China \\
{\tt\small chuanrui001@e.ntu.edu.sg, ziwei.wang@ntu.edu.sg} \\
\href{https://xingyoujun.github.io/rodyn/}{\tt\small https://xingyoujun.github.io/rodyn/}
}

\begin{document}


\twocolumn[{%
\renewcommand\twocolumn[1][]{#1}%

\maketitle

\vspace{-0.6cm}
\begin{center}
    \centering
    \captionsetup{type=figure}
    \includegraphics[width=\textwidth]{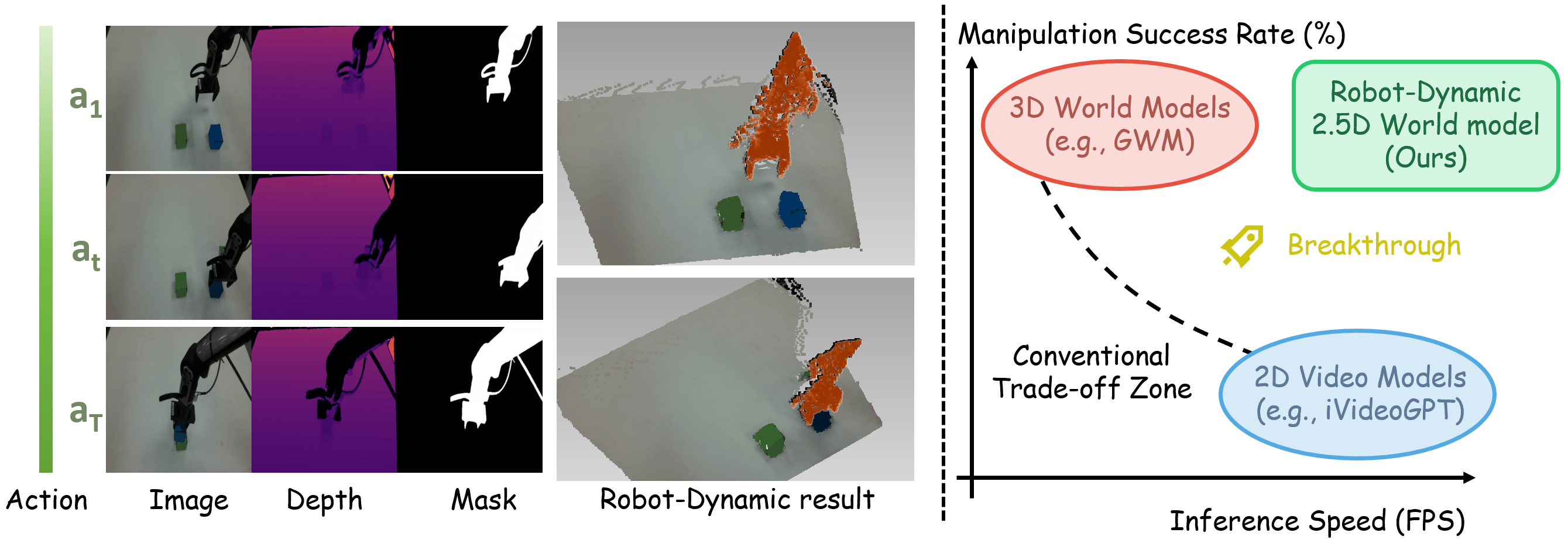}
    \vspace{-0.4cm}
    \captionof{figure}{
    RoDyn achieves a new state-of-the-art trade-off in learned world models. 
    In the left image RoDyn generates physically-grounded future 2.5D representations (Image, Depth, Mask) conditioned on robot actions. 
    }
    \vspace{-0.2cm}
    \label{fig:teaser}
\end{center}%
}]

\begin{abstract}
Learned world models hold significant potential as neural simulators for robotic manipulation. 
However, prevalent 2D video-based models inherently lack the spatial and kinematic reasoning crucial for physical interactions. 
We introduce RoDyn, a novel Robot-Dynamic 2.5D World Model that formulates environmental dynamics within a highly efficient, geometry-aware latent space. 
Through the proposed Robot-Dynamic Tokenizer, we explicitly couple semantic visual appearances with spatial and agent-centric priors via an RGB-dominated cross-attention mechanism and dynamic mask guidance. Furthermore, by injecting these mask priors directly into sequence transitions, our Mask-guided Autoregressive architecture drives the model to focus on active robot-object interaction regions.
Extensive experiments demonstrate that RoDyn establishes SOTA generation fidelity across large-scale datasets. 
Crucially, it translates these predictive capabilities into substantial downstream gains, accelerating model-based reinforcement learning and achieving a 42\% improvement in real-world imitation learning success rates over pure 2D baselines. 
\end{abstract}

\section{INTRODUCTION}
In recent years, learned world models~\cite{bar2025navigation} have emerged as a powerful paradigm, driving substantial advances across diverse domains such as scene generation~\cite{liang2025wonderland, zou2025mudg, zhang2025transplat}.
By leveraging limited observations to forecast future states, these models demonstrate a remarkable ability to simulate environmental dynamics and facilitate adaptation to novel scenarios. 
These predictive capabilities make them particularly valuable for robotic manipulation~\cite{lu2025gwm, zhang2024category, zhang2026unipr}, where accurately anticipating physical interactions is essential for reliable planning and execution. 
While many recent models~\cite{zhen2025tesseract} excel at generating high-fidelity videos using text instructions or goal-conditioning~\cite{hong2022cogvideo, wan2025wan}, their utility in embodied AI is fundamentally restricted by a lack of explicit action integration. 
To act as effective neural simulators, world models must incorporate action-based conditioning~\cite{wu2024ivideogpt}, grounding the generative process in physical control to ensure that predicted future states directly correspond to the causal consequences of embodied decision-making.

Despite recent advancements in action conditioning, existing models~\cite{wu2024ivideogpt, guo2025dynamical} remain fundamentally constrained to 2D RGB pixel prediction and lack explicit spatial representations.
This limitation is particularly detrimental to robotic manipulation, where continuous physical interactions inherently unfold in three-dimensional space. 
Without robust geometric understanding, purely video-based world models are prone to hallucinating physically implausible dynamics, especially under severe occlusions or complex object manipulations.
In contrast, recent efforts such as GWM~\cite{lu2025gwm}, which integrate 3D Gaussian Splatting~\cite{kerbl3Dgaussians}, attempt to alleviate this issue by incorporating explicit 3D structure.
However, their performance depends heavily on the quality of dense 3D reconstruction, making them brittle in monocular observation settings and difficult to scale across diverse scenes. 
More critically, explicit 4D world modeling incurs substantial computational and memory overhead, resulting in inference speeds that are insufficient for high-frequency robotic control.
Consequently, achieving the spatial awareness of 3D models while preserving the computational efficiency of video-based approaches remains a critical open challenge.

To address the aforementioned limitations, we propose \textbf{RoDyn}, a novel \textbf{Ro}bot-\textbf{Dyn}amic 2.5D World Model.
As illustrated in \Cref{fig:teaser}, RoDyn aims to combine the fast inference speed of 2D video models with the spatial awareness and high manipulation success rates typically associated with heavy 3D world models.
Rather than relying on computationally prohibitive dense 3D reconstructions or physically agnostic 2D pixels, RoDyn seamlessly integrates visual context with geometric and physical constraints through a unified network.
We first introduce a novel Robot-Dynamic Tokenizer to efficiently construct a geometry-aware 2.5D latent space. 
Instead of employing shallow multimodal fusion, this module leverages an RGB-dominated cross-attention mechanism with dynamic mask guidance, in which spatial depth and robot priors explicitly modulate core visual features. 
This design enables the tokenizer to adaptively concentrate its representational capacity on active robot–object interactions.
Building upon these structured tokens, we further design a mask-guided autoregressive (AR) Transformer for future state prediction. 
By injecting mask priors into sequence transitions, the AR model localizes temporal dynamics within the robot’s active workspace and suppresses attention to non-masked static regions.
Ultimately, this coupled formulation enables RoDyn to achieve explicit 3D spatial awareness while preserving the inference efficiency of 2D models.

Empirically, our method achieves best performance in both visual generation quality and spatial accuracy across large-scale benchmarks. 
Most notably, in real-world manipulation deployments, policies trained on RoDyn-synthesized rollouts converge substantially faster and improve overall success rates by more than 42\% compared to pure 2D baselines, effectively matching the ground-truth data.
Comprehensively, our main contributions are summarized as follows:
\begin{itemize}
\item[$\bullet$] 
We propose \textbf{RoDyn}, a novel Robot-Dynamic 2.5D World Model that bridges the critical gap between 2D computational efficiency and 3D spatial awareness by formulating environmental dynamics.

\item[$\bullet$] 
We introduce a unified geometry-aware architecture comprising Robot-Dynamic Tokenizer and Mask-guided Autoregressive Transformer. 


\item[$\bullet$] 
Extensive experiments demonstrate that RoDyn achieves best visual and spatial performance and improves success rates in real-world manipulation tasks by more than 42\% compared to 2D baselines.
\end{itemize}

\section{Related Work}

\subsection{Learned World Models}
Learning to predict future states of the observed world from action-based control inputs has long been a central challenge in domains such as autonomous driving~\cite{zuo2025gaussianworld, wang2024drivedreamer}, game agents~\cite{cheng2025playing}, and robotic manipulation~\cite{seo2023masked, lu2025gwm}. 
Early approaches~\cite{wu2024ivideogpt} are inspired by VideoGPT, where RGB video frames are tokenized and future color images are generated in an autoregressive manner. 
More recently, advances in diffusion-based video generation~\cite{kondratyuk2023videopoet, wan2025wan} have motivated world models~\cite{liang2025wonderland, wu2025cat4d} to predict future sequences within video latent spaces. 
Meanwhile, GWM~\cite{lu2025gwm} adopts 3DGS as its underlying representation, however, limited pretraining data and the insufficient quality of 3DGS in monocular settings severely restrict its generative capacity on public datasets and real-world scenes. 
Nevertheless, improving 3D representation remains a critical prerequisite for advancing the applicability of learned world models to robotic manipulation.

\subsection{4D Video Prediction}
4D video generation~\cite{yin20234dgen, singer2023text} has attracted considerable attention in recent years, driven by advances in diffusion models and 3DGS techniques. 
Previous approaches~\cite{lu2025gwm} utilize 3DGS to reconstruct the 3D geometry of generated videos, but these methods typically require multi-view consistent video generation, which limits their applicability to robotic manipulation. 
More recently, TesserAct~\cite{zhen2025tesseract}explores RGB-DN video generation, highlighting the importance of monocular 3D information for embodied tasks. 
Nonetheless, these approaches lack explicit action conditioning, which is essential for enhancing robotic manipulation performance.

\subsection{World Model for Robotic Manipulation}
The primary objective of incorporating world models into robotic manipulation is to enrich real-world data for imitation learning in unseen scenarios~\cite{kim2024openvla,li2025controlvla} and to serve as simulators for reinforcement learning (RL)~\cite{luo2024serl, luo2025precise} .
Previous approaches~\cite{lu2025gwm} have integrated world model observation encoders as feature extractors for imitation learning policies, which can largely be regarded as pretraining strategies. 
Other methods~\cite{wu2024ivideogpt} employ world models to support RL by reducing the high cost associated with constructing digital twins of the required environments. 
However, these approaches generally lack explicit 3D information, which limits their effectiveness both as simulators and as generators of realistic training data for real-world applications.

\section{Method}

\begin{figure*}[tb]
  \centering
  \includegraphics[width=1\textwidth]{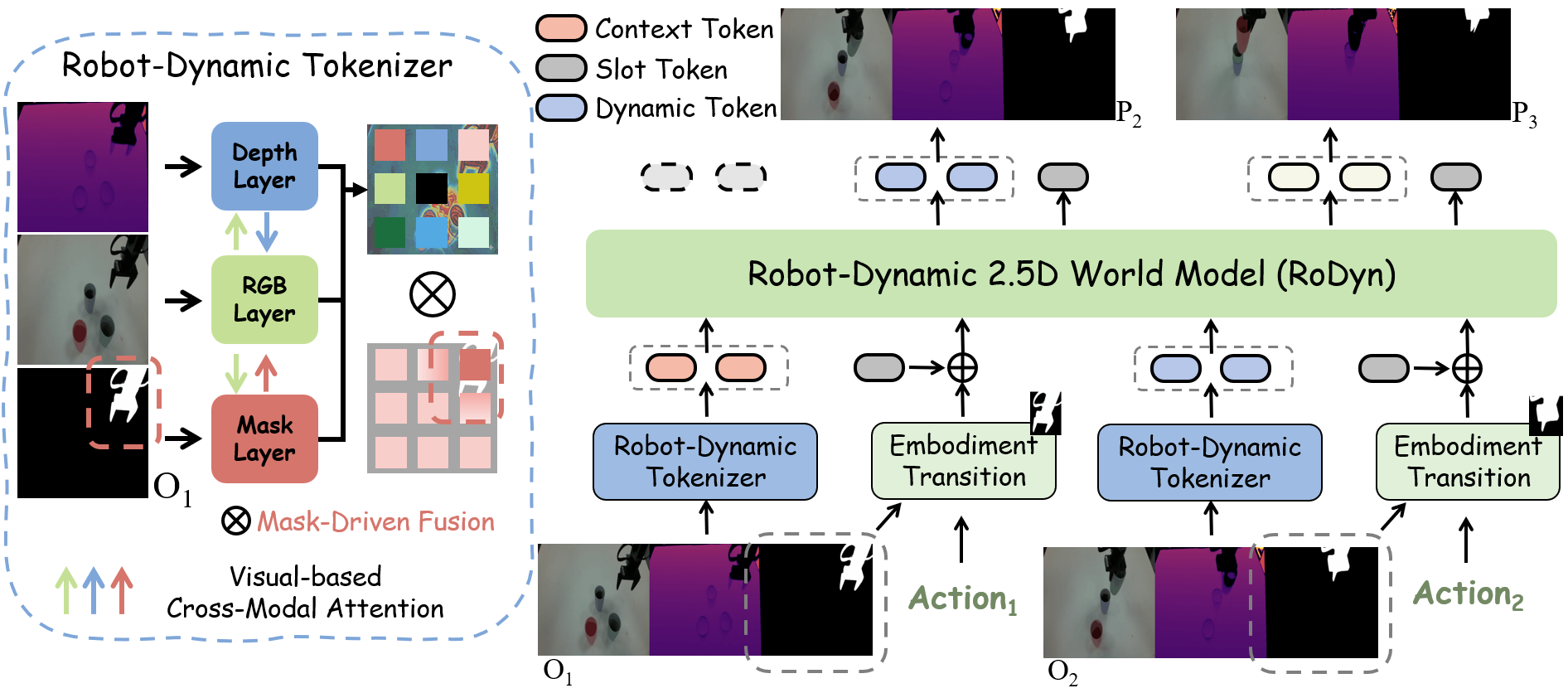}
  \vspace{-0.3cm}
\caption{
  \textbf{Framework of the proposed RoDyn.} 
  Multi-modal 2.5D inputs are encoded into physics-aware discrete tokens via the Robot-Dynamic Tokenizer, which features RGB-dominated cross-attention and mask-driven fusion.
  To enforce causal physical dynamics during sequence transitions, both robotic action trajectories and extracted kinematic masks are explicitly injected into the Embodiment Transition Tokens. 
  The Mask-guided Autoregressive Transformer then sequentially predicts future tokens ($\mathbf{P}_t$), strictly grounding the generation on prior context and active robot-object interactions.
}

  \label{fig:pipeline}
  \vspace{-0.4cm}
\end{figure*}

We introduce \textbf{RoDyn}, a novel Robot-Dynamic 2.5D World Model that autoregressively predicts future physical states through the joint synthesis of semantic visual appearances, spatial depth, and dynamic robot masks. 
An overview of the proposed framework is illustrated in \Cref{fig:pipeline}.
Given an initial RGB observation $\mathbf{I}_1$, we construct a geometry-aware 2.5D observation space $\mathbf{O}_1$. 
Our novel Robot-Dynamic Tokenizer then processes this input via an RGB-dominated cross-attention mechanism, encoding $\mathbf{O}_1$ into a unified, physics-aware discrete token representation $\mathbf{Z}_1$ (\cref{sec:mmtokenizer}).
Conditioned on the continuous action trajectory $\{\mathbf{a}_t\}_{t=1}^{T}$, a Mask-guided Autoregressive Transformer sequentially predicts future tokens (\cref{sec:gpt}). 
Crucially, this module injects mask priors directly into sequence transitions to focus on causal robot-environment interactions.
Finally, we demonstrate that this physically constrained formulation allows RoDyn to serve as a highly efficient neural simulator, translating explicit spatial awareness into substantial gains for downstream tasks, including real-world imitation learning and model-based reinforcement learning (\cref{sec:tasks}).

\subsection{Robot-Dynamic Tokenizer}
\label{sec:mmtokenizer}
Conventional tokenizers~\cite{esser2021taming} are predominantly designed for physically agnostic 2D pixels, limiting their applicability in 3D-aware robotic tasks. 
To project our 2.5D state space into a compact, physics-aware discrete representation, we propose the Robot-Dynamic Tokenizer, instantiated within a dual-encoder--decoder framework $\{(\mathcal{E}_c,\mathcal{D}_c), (\mathcal{E}_p,\mathcal{D}_p)\}$~\cite{wu2024ivideogpt}.
To explicitly align multi-modal inputs without relying on naive channel-wise concatenation, we employ an RGB-dominated cross-attention mechanism. 
After parallel processing in modality-specific layers, spatial depth and agent-centric masks explicitly query and modulate the core semantic visual features. 
Subsequently, a mask-driven dynamic fusion operation ($\otimes$) injects these mask priors to adaptively concentrate the representation on active robot-object interactions. 
This tight, geometry-aware coupling guarantees that the resulting tokens are strictly grounded in causal robotic dynamics rather than static background noise.

Through this carefully designed architecture, the Robot-Dynamic Tokenizer generates unified, geometry-aware tokens. We first employ the context encoder-decoder pair $(\mathcal{E}_c,\mathcal{D}_c)$ to extract contextual tokens from the initial observations $\mathbf{O}_{1:T_0}$, establishing the global static scene:
\begin{equation}
\mathbf{Z}_t^c = \mathcal{E}_c(\mathbf{O}_t), \quad
\hat{\mathbf{O}}_t = \mathcal{D}_c(\mathbf{Z}_t^c) \qquad t=1,\dots,T_0.
\end{equation}

In contrast, future frames exhibit substantial temporal redundancy, with significant changes strictly localized to the interacting entities. 
To exploit this physical property, the conditional encoder-decoder pair $(\mathcal{E}_p,\mathcal{D}_p)$ is dedicated exclusively to modeling temporal evolution. Guided by the aforementioned mask-driven fusion, this predictive module selectively generates tokens for dynamic regions:
\begin{equation}
\begin{aligned}
\mathbf{Z}_t^d &= \mathcal{E}_p(\mathbf{O}_t|\mathbf{O}_{1:T_0}), \\
\hat{\mathbf{O}}_t &= \mathcal{D}_p(\mathbf{Z}_t^d|\mathbf{O}_{1:T_0}),
\qquad t = T_0+1,\dots,T.
\end{aligned}
\end{equation}
This conditioning is implemented via cross-attention between the historical context tokens and future observations. 
Consequently, our tokenizer drastically reduces computational redundancy while preserving exquisite multi-modal detail in the active physical interaction zones.

To optimize the Robot-Dynamic Tokenizer, we employ a joint training objective designed to ensure both high-fidelity multi-modal reconstruction and a stable discrete latent space. 
The overall loss function is formulated as:
\begin{equation}
\begin{aligned}
\mathcal{L}_\mathrm{tokenizer} &= \sum_{t=1}^{T_0}\mathcal{L}_\mathrm{VQGAN}(\mathbf{O}_t;\mathcal{E}_c,\mathcal{D}_c) \\
&+ \sum_{t=T_0+1}^{T}\mathcal{L}_\mathrm{VQGAN}(\mathbf{O}_t;\mathcal{E}_p,\mathcal{D}_p),
\end{aligned}
\end{equation}
where $\mathcal{L}_\mathrm{VQGAN}$ comprises an $\mathcal{L}_1$ reconstruction loss, a commitment loss~\cite{van2017neural}, a perceptual loss~\cite{johnson2016perceptual}, and optionally an adversarial loss~\cite{esser2021taming}.


\subsection{Mask-guided Autoregressive Transformer}
\label{sec:gpt}

Following the extraction of context and dynamic tokens by our Robot-Dynamic Tokenizer, we formulate future state generation as a discrete sequence modeling problem. The high-dimensional 2.5D multi-modal inputs are serialized into a unified token sequence: $\mathbf{X}_T = (\mathbf{Z}_1^c, [\mathbf{S}_1], \dots, \mathbf{Z}_{T_0}^c, [\mathbf{S}_{T_0}], \mathbf{Z}_{T_0+1}^d, [\mathbf{S}_{T_0+1}], \dots, \mathbf{Z}_{T}^d)$.

As illustrated in \Cref{fig:pipeline}, to strictly enforce causal constraints during sequence transitions, we introduce Embodiment Transition Token, denoted as $[\mathbf{S}_t]$. Rather than relying solely on simplistic action conditioning, our transition token acts as a rigorous physical boundary flag that explicitly fuses the intended control command with current kinematic priors. Specifically, we inject both the continuous action trajectory and the mask-guided spatial layout into the latent space:
\begin{equation}
[\mathbf{S}_t] = [\mathbf{S}_\mathrm{base}] + \mathcal{P}_\mathrm{act}(\mathbf{a}_t) + \mathcal{P}_\mathrm{mask}(\mathbf{m}_t),
\end{equation}
where $[\mathbf{S}_\mathrm{base}]$ represents a learnable universal boundary embedding. $\mathcal{P}_\mathrm{act}(\cdot)$ is a kinematic projection network that encodes the end-effector pose $\mathbf{a}_t$ (including position, orientation, and gripper state). Crucially, $\mathcal{P}_\mathrm{mask}(\cdot)$ injects the extracted physical mask representation $\mathbf{m}_t$. 
This explicit mask-guided injection drives the autoregressive model to ground its temporal predictions strictly on active robot-object interactions, significantly mitigating physical hallucinations.

Operating on this unified and mask-guided sequence, we deploy a highly scalable causal transformer backbone, adopting a LLaMA-style architecture~\cite{touvron2023llama} equipped with RMSNorm~\cite{zhang2019root}, SwiGLU activations~\cite{shazeer2020glu}, and Rotary Positional Embeddings (RoPE)~\cite{su2024roformer}. To maximize computational efficiency and force the network's capacity toward learning critical physical dynamics, the predictive objective is exclusively applied to the dynamic tokens. The context tokens function strictly as read-only prefix prompts, and the model is optimized via a spatially-selective cross-entropy loss:
\begin{equation}
\mathcal{L}_\mathrm{AR} = -\sum_{t=T_0+1}^{T} \log P(\mathbf{Z}_t^d \mid \mathbf{X}_{<t}),
\end{equation}
where $\mathbf{X}_{<t}$ denotes the history of all serialized tokens up to the current transition boundary. 
To ensure uniform batch processing and consistent alignment of contextual regions, we pad the sequences with empty tokens $[\mathbf{E}]$. 
During inference, future dynamic states are generated in a strictly autoregressive manner, with newly predicted spatio-kinematic tokens fed back into the sequence to simulate continuous, closed-loop robotic manipulation rollouts.

\subsection{Downstream Tasks}
\label{sec:tasks}
\noindent \textbf{RoDyn for Imitation Learning.}
By predicting future observations conditioned on continuous actions, world models serve as a powerful data augmentation engine for real-world robotic tasks. 
In our framework, we employ RoDyn to synthesize multiple geometrically grounded demonstration rollouts given an initial observation and a successful action trajectory. 
The generated 2.5D multi-modal outputs are subsequently utilized to train a robust robotic manipulation policy via 3D Diffusion Policy~\cite{chi2023diffusion}. 
For our imitation learning setup, the policy ingests only the synthesized semantic appearances and spatial topologies, while actions are predicted in sequential chunks to ensure temporal consistency during continuous robotic control.

\noindent \textbf{RoDyn for Reinforcement Learning.}
Similar to recent visual world models~\cite{wu2024ivideogpt, lu2025gwm}, RoDyn functions as a highly stable, physically grounded neural simulator within existing model-based RL frameworks. 
The objective of model-based RL~\cite{janner2019trust} is to learn a policy $\pi$ that maximizes the cumulative reward $\mathbf{r}$ through policy roll-outs. We formalize this as a Markov Decision Process (MDP) defined by $(\mathbf{S}, \mathbf{A}, \mathcal{P}, \mathbf{r}, \gamma)$, where $\mathbf{S}$ and $\mathbf{A}$ denote the state and action spaces, $\mathcal{P}$ represents the world model that predicts physical state transitions, and $\gamma$ is the discount factor. 
We first construct a real replay buffer $\mathcal{D}_\mathrm{real}$ using an initialized policy $\pi$ and train RoDyn $\mathcal{P}_{\theta}$ on $\mathcal{D}_\mathrm{real}$. Subsequently, the optimized $\mathcal{P}_{\theta}$ is employed to generate an imagined replay buffer $\mathcal{D}_\mathrm{imag}$ under identical action roll-outs. The actor–critic model $(\pi, v)$ is then updated using the combined dataset $\mathcal{D}_\mathrm{real} \cup \mathcal{D}_\mathrm{imag}$. Furthermore, we incorporate an auxiliary reward head into RoDyn to predict the step-wise reward for each action step, consistent with standard visual RL practices~\cite{wu2024ivideogpt}.

\section{Experiments}

\begin{figure*}[tb]
  \centering
  \includegraphics[width=1\linewidth]{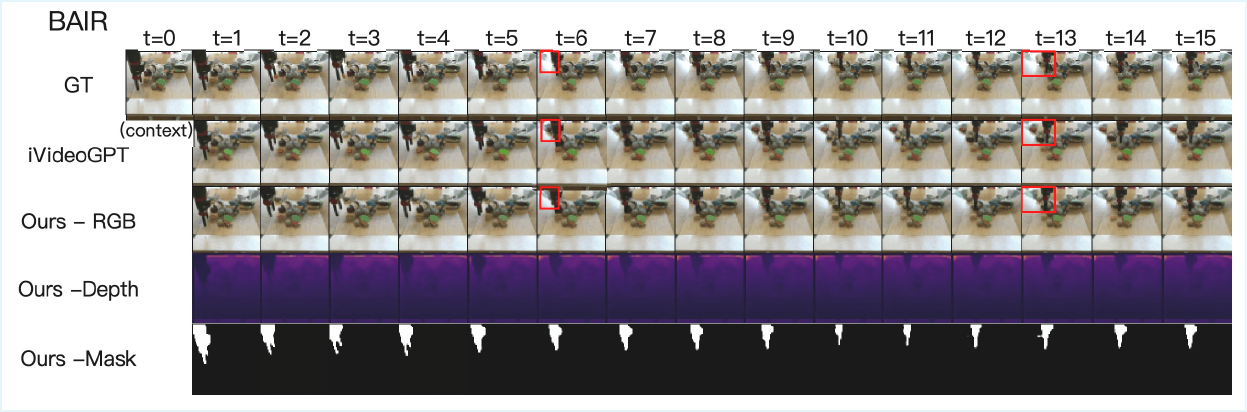}
  \caption{
  \textbf{Comparisons on BAIR dataset.}
  We present action-conditioned multi-modal video generation results on the BAIR dataset, comparing with iVideoGPT. 
  Our method produces better generations and captures richer geometric information.
  }
  \label{fig:bair}
  \vspace{-0.4cm}
\end{figure*}

In this section, we comprehensively evaluate \textbf{RoDyn} against prior state-of-the-art methods across diverse simulated and real-world robotic manipulation tasks.
Given the embodied nature of our approach, we first deploy RoDyn in physical environments, demonstrating its superior multi-modal generation fidelity and substantial improvements in real-world manipulation success rates (\cref{sec:rwil}). 
Next, to rigorously assess its zero-shot generalization capabilities, we benchmark its action-conditioned generation quality across large-scale open-source datasets, including the challenging DROID dataset (\cref{sec:acmvg}). 
We then evaluate its efficacy as a physically grounded neural simulator for accelerating visual model-based reinforcement learning (\cref{sec:vmrl}). 
Finally, we conduct extensive ablation studies to analyze the specific contributions of the Robot-Dynamic Tokenizer and the mask-guided AR (\cref{sec:ab}). 

\subsection{Real-World Imitation Learning}
\label{sec:rwil}
To rigorously evaluate the practical utility and physical grounding of our generated roll-outs, we conduct extensive real-world imitation learning experiments. 
The physical deployment utilizes a GALAXEA A1 robotic arm, with a third-person Intel RealSense D435i camera capturing high-fidelity RGB-D observations. 
We train RoDyn across 10 distinct robotic manipulation tasks, collecting 50 human demonstrations per task. 
For hardware evaluation, we select 5 representative tasks from our dataset, encompassing complex spatial interactions such as multi-stage item stacking and precise pick-and-place operations under varied spatial distractors. 
Using the pre-trained RoDyn (further detailed in \cref{sec:acmvg}), we synthesize imagined execution trajectories conditioned on the original demonstration actions. 
These synthesized 2.5D multi-modal outputs are then utilized to train a robust 3D Diffusion Policy~\cite{chi2023diffusion}. 

\begin{figure}[tb]
  \centering
  \includegraphics[width=1\linewidth]{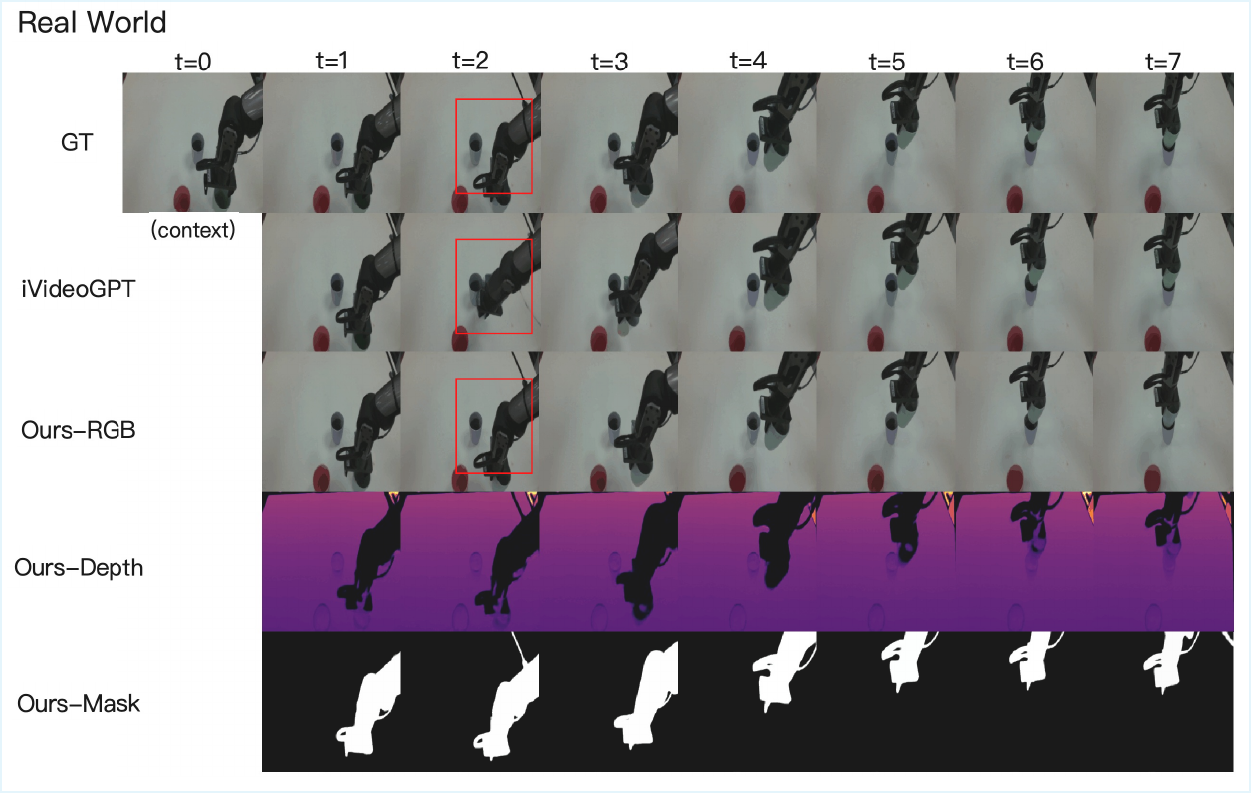}
  \vspace{-0.4cm}
  \caption{
  \textbf{Comparison on real data.}
  We showcase qualitative results on real-world, high-resolution data. 
  }
  \label{fig:real_compare}
  \vspace{-0.6cm}
\end{figure}

\begin{figure}[tb]
  \centering
  \includegraphics[width=1\linewidth]{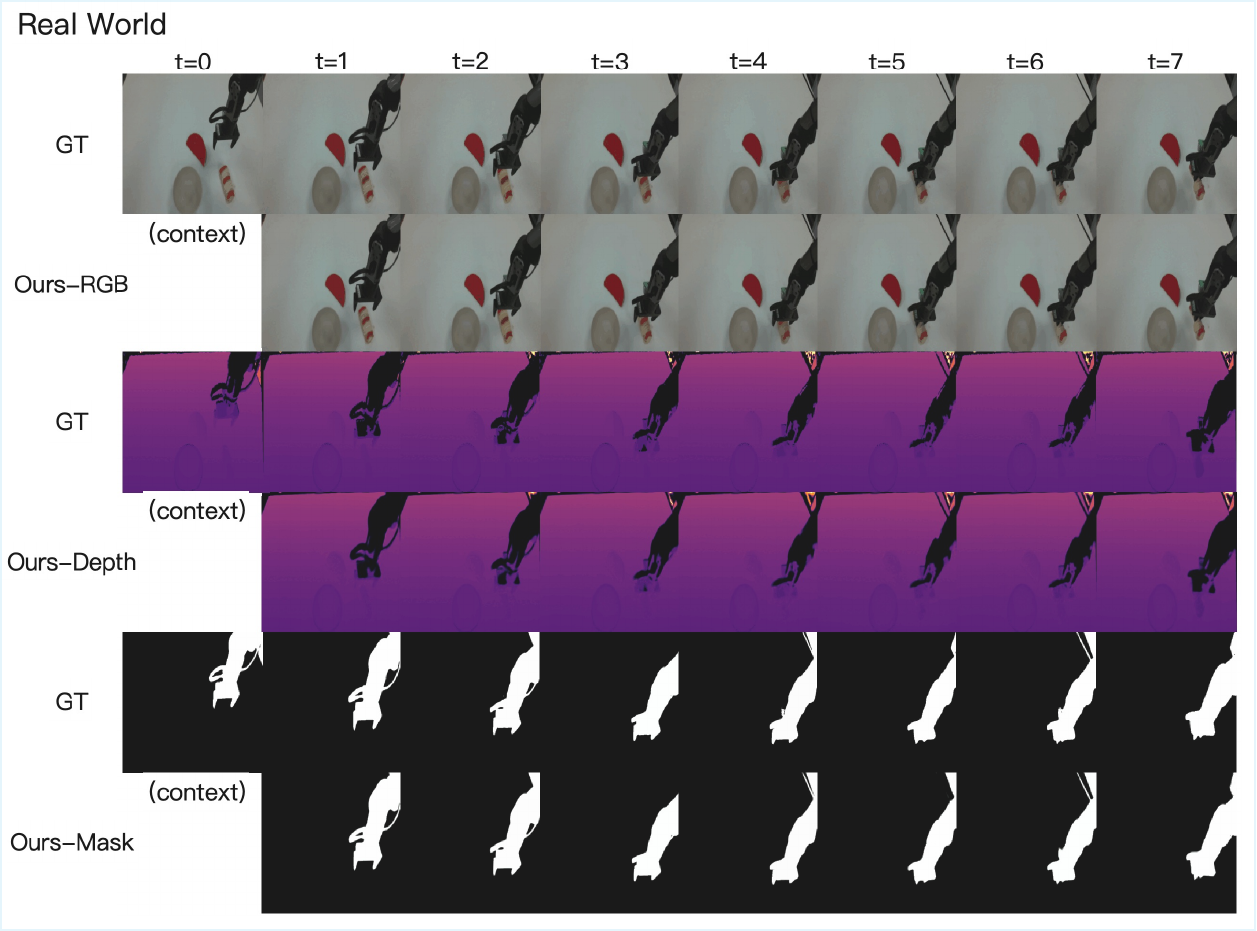}
  \caption{
  \textbf{Qualitative Result on Real Data.}
  We present qualitative results on real-world, high-resolution data.
  }
  \label{fig:vis_real}
\end{figure}

To isolate the contribution of our geometry-aware 2.5D representation, we compare the policies trained on RoDyn-generated data against those trained on a pure 2D baseline (iVideoGPT, augmented with ground-truth depth to ensure a fair comparison) as well as the original real-world demonstrations (GT). 
The quantitative results are presented in \Cref{tab:deploy}. Across the 5 diverse deployment tasks, RoDyn achieves an average manipulation success rate of 87.0\%, outperforming the 2D baseline by an absolute margin of 42.0\%. Notably, our method matches and slightly surpasses the performance of policies trained entirely on ground-truth data (83.0\%). 
This indicates that the geometrically consistent and mask-guided roll-outs produced by RoDyn help bridge the reality gap, functioning as a data engine that smoothes sub-optimal human demonstration noise for real-world robotic control.
Furthermore, the quantitative visual fidelity results in \Cref{tab:realvis} show that RoDyn performs well on high-resolution inputs, as richer 3D structural and kinematic priors are explicitly incorporated. 
Qualitative comparisons on real-world, high-resolution data are further presented in \Cref{fig:real_compare} and \Cref{fig:vis_real}.

\begin{table}[tb]
\centering
\caption{\textbf{Real-world Manipulation Deployment Results.} 
We evaluate the policies trained on synthesized data across five diverse real-world tasks. The table reports the success rate for each task. Our RoDyn model significantly outperforms the 2D baseline (iVideoGPT) and achieves a performance level that effectively matches policies trained directly on real-world ground-truth (GT) demonstrations.}
\begin{tabular}{l c c c}
\toprule
\textbf{GALAXEA A1 Task} & iVideoGPT & \textbf{RoDyn (Ours)} & GT \\
\midrule
\multicolumn{4}{c}{\textit{Stacking Tasks}} \\
\midrule
Stack Green Cup       & 50\% & \cellcolor{green!14}\textbf{90\%} & \cellcolor{red!20}90\% \\
Stack Two Cups        & 30\% & \cellcolor{green!14}\textbf{70\%} & \cellcolor{red!20}60\% \\
Stack Cubes           & 45\% & \cellcolor{green!14}\textbf{90\%} & \cellcolor{red!20}85\% \\
\midrule
\multicolumn{4}{c}{\textit{Pick and Place Tasks}} \\
\midrule
Pick and Place Banana & 55\% & \cellcolor{green!14}\textbf{95\%} & \cellcolor{red!20}95\% \\
Pick and Place Bread  & 45\% & \cellcolor{green!14}\textbf{90\%} & \cellcolor{red!20}85\% \\
\midrule
\textbf{Total Success} & 45\% & \cellcolor{green!14}\textbf{87\%} & \cellcolor{red!20}83\% \\
\bottomrule
\end{tabular}
\label{tab:deploy}
\vspace{-0.4cm}
\end{table}

\begin{table}[tb]
\centering
\caption{\textbf{Quantitative comparisons on Real-World Data.} 
We benchmark our method against iVideoGPT, on high-resolution ($256 \times 256$) real-world manipulation sequences. 
}
\begin{tabular}{lcccc}
\toprule
\textbf{Real-world} & PSNR$\uparrow$ & SSIM$\uparrow$ & LPIPS$\downarrow$ \\
\midrule
iVideoGPT~\cite{wu2024ivideogpt} & 32.14 & 0.901 & 0.130 \\
\textbf{Our} & \textbf{38.11} & \textbf{0.941} & \textbf{0.091}\\
\bottomrule
\end{tabular}
\label{tab:realvis}
\vspace{-0.2cm}
\end{table}

\begin{table}[tb]
\centering
\caption{\textbf{Quantitative comparisons on BAIR and RoboNet dataset.}
 (\textbf{Bold} numbers indicate the best, and \underline{underlined} numbers indicate the second-best.)
}
\begin{tabular}{lccccc}
\toprule
\textbf{BAIR}~\cite{ebert2017self} & FVD$\downarrow$ & PSNR$\uparrow$ & SSIM$\uparrow$ & LPIPS$\downarrow$ & AbsRel$\downarrow$  \\
\midrule
MaskViT~\cite{gupta2022maskvit} & 70.5 & - & - & - & - \\
iVideoGPT~\cite{wu2024ivideogpt} & \underline{65.01} & \underline{23.40} & \underline{0.882} & \underline{0.058} & \underline{0.059} \\
\textbf{Our} & \textbf{60.9} & \textbf{23.82} & \textbf{0.896} & \textbf{0.051} & \textbf{0.045} \\
\toprule
\textbf{RoboNet}~\cite{dasari2019robonet} & FVD$\downarrow$ & PSNR$\uparrow$ & SSIM$\uparrow$ & LPIPS$\downarrow$ & AbsRel$\downarrow$  \\
\midrule
MaskViT~\cite{gupta2022maskvit} & 133.5 & 23.2 & 0.805 & 0.042 & - \\
SVG~\cite{villegas2019high} & 123.2 & 23.9 & 0.878 & 0.060 & - \\
GHVAE~\cite{wu2021greedy} & 95.2 & 24.7 & 0.891 & \underline{0.036} & - \\
iVideoGPT~\cite{wu2024ivideogpt} & \textbf{65.8} & \underline{27.2} & \underline{0.895} & 0.053 & \underline{0.054} \\
\textbf{Our} & \underline{67.6} & \textbf{28.0} & \textbf{0.920} & \textbf{0.032} & \textbf{0.031} \\
\bottomrule
\end{tabular}
\label{tab:bair}
\end{table}

\begin{table}[tb]
\centering
\caption{\textbf{Quantitative Evaluation on the DROID Dataset.} We compare our proposed RoDyn against recent state-of-the-art video generation models.}
\begin{tabular}{lcccc}
\toprule
\textbf{Methods} & PSNR $\uparrow$ & SSIM $\uparrow$ & LPIPS $\downarrow$ & FVD $\downarrow$ \\
\midrule
IRASim~\cite{zhu2025irasim}       & 22.11 & 0.846 & 0.119 & 175.7 \\
Cosmos~\cite{agarwal2025cosmos}   & 21.13 & 0.826 & 0.122 & 184.3 \\
EVAC~\cite{li2025evac}         & 21.97 & 0.877 & 0.124 & 219.8 \\
BridgeV2W~\cite{chen2026bridgev2w}& 22.89 & 0.874 & 0.111 & 145.2 \\
\midrule
\textbf{RoDyn (Ours)}             & \textbf{24.37} & \textbf{0.877} & \textbf{0.076} & \textbf{122.7} \\
\bottomrule
\end{tabular}
\label{tab:droid_results}
\end{table}

\begin{figure*}[tb]
  \centering
  \includegraphics[width=0.9\linewidth]{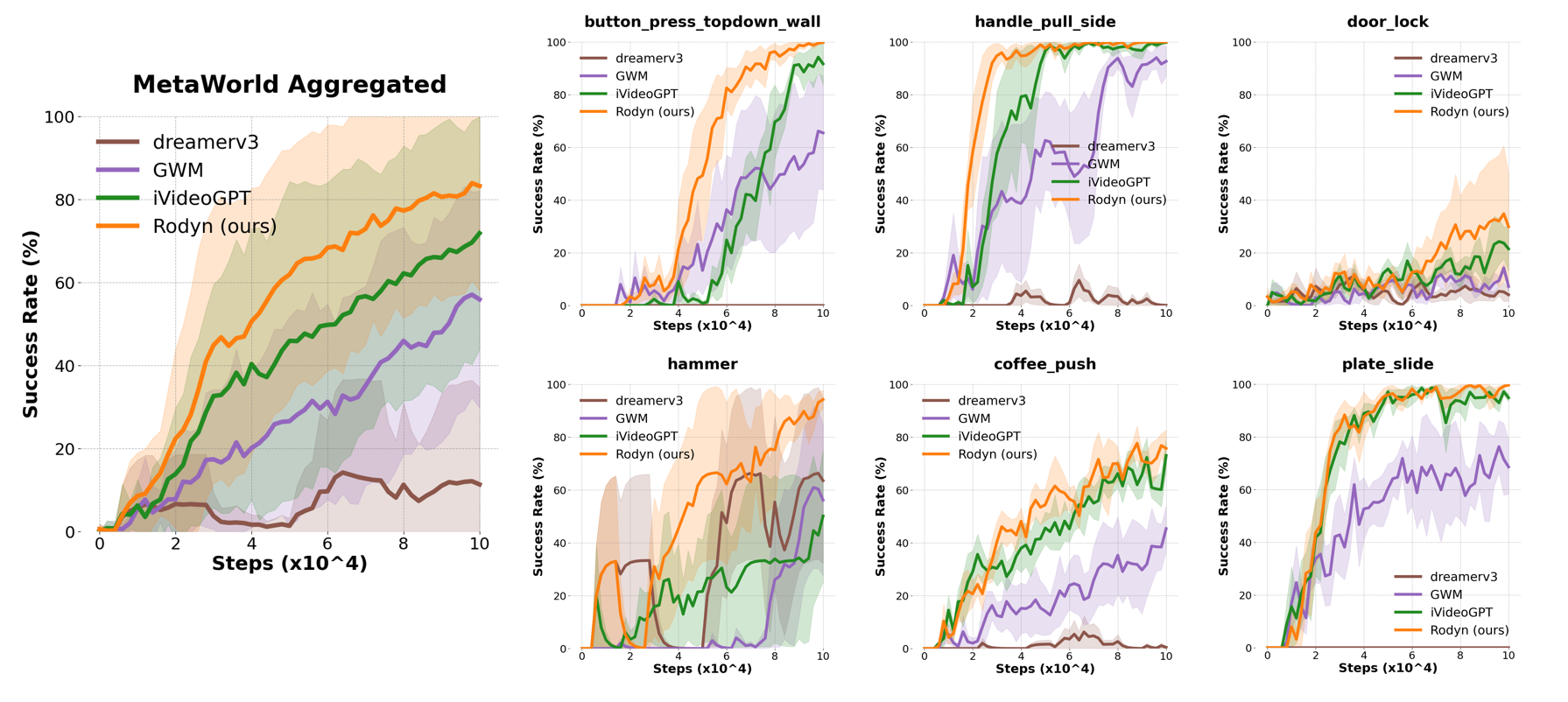}
  \caption{
  \textbf{MBRL Results on Metaworld.}
  The shaded region indicates the 95\% confidence interval (CI) across three random seeds, with each data point evaluated over 20 episodes. 
  }
  \label{fig:mbrl}
  \vspace{-0.3cm}
\end{figure*}

\subsection{Action-conditioned Multi-modal Video Generation}
\label{sec:acmvg}

To rigorously evaluate the zero-shot generalization and physically grounded generation capabilities of RoDyn, we benchmark it on the action-conditioned multi-modal video generation task across multiple large-scale datasets. We utilize the BAIR robot pushing dataset~\cite{ebert2017self} (43k training and 256 testing videos) and a single-arm subset of the RoboNet dataset~\cite{dasari2019robonet} (19k training and 256 testing videos) to manage computational overhead while preserving environmental complexity. Crucially, to assess the model's scalability and robustness to extreme real-world diversity, we incorporate the highly challenging DROID dataset. Following the experimental protocol established by Chen et al.~\cite{chen2026bridgev2w} (19k training 200 testing).
To construct our geometry-aware 2.5D state space, we extract metric depth maps using Video Depth Anything~\cite{chen2025video} and generate temporally consistent, agent-centric kinematic masks utilizing Grounding DINO~\cite{liu2024grounding} combined with SAM~2~\cite{ravi2024sam}. While the open-source datasets are processed at a $64 \times 64$ resolution, we additionally employ our self-collected dataset (1k training and 150 testing videos spanning six complex manipulation tasks) at a high resolution of $256 \times 256$ to evaluate fine-grained spatial and visual fidelity.
To quantify visual quality and semantic preservation, we report FVD~\cite{unterthiner2018towards}, PSNR~\cite{huynh2008scope}, SSIM~\cite{wang2004image}, and LPIPS~\cite{zhang2018unreasonable}. 
For spatial topology accuracy, we adopt the AbsRel (Absolute Relative Error: $\tfrac{|\hat{d} - d|}{d}$)~\cite{yang2024depth} metric for depth predictions. 
To ensure a rigorous and fair evaluation, we generate four separate stochastic roll-outs for both iVideoGPT and RoDyn. 
For pure 2D baselines lacking explicit 3D awareness, we apply Video Depth Anything to their synthesized RGB roll-outs to extract depth maps for AbsRel computation.
The quantitative results across the BAIR, RoboNet, and DROID datasets are presented in \Cref{tab:bair} and \Cref{tab:droid_results}. 
On both the BAIR and DROID datasets, RoDyn significantly outperforms all existing baselines across all visual and spatial metrics. 
Qualitative comparisons on the BAIR dataset are shown in \Cref{fig:bair}, where RoDyn produces more accurate generations and captures richer geometric information than iVideoGPT.
On the RoboNet dataset, RoDyn achieves the best performance in pixel-level fidelity (PSNR, SSIM, LPIPS) and spatial accuracy (AbsRel). 

\subsection{Visual Model-Based Reinforcement Learning}
\label{sec:vmrl}
To evaluate the efficacy of RoDyn as a physically grounded neural simulator, we conduct visual model-based Reinforcement Learning (RL) experiments on six challenging robotic manipulation tasks from the Meta-World benchmark~\cite{yu2020meta}. 
We employ a framework inspired by MBPO~\cite{janner2019trust}, utilizing DrQ-v2~\cite{yarats2021mastering} as the foundational actor-critic algorithm. To ensure a rigorous and fair evaluation, we benchmark against state-of-the-art methods across different paradigms, including explicit 3D-aware world models (GWM) and highly optimized pure 2D models (iVideoGPT, initialized from pre-trained checkpoints, and DreamerV3).
The learning curves in \Cref{fig:mbrl} demonstrate the consistent improvements of our approach. 
As shown in the aggregated results, RoDyn elevate the average success rate by an absolute margin of over 10\% compared to the strongest baseline. 
Crucially, this performance leap is not an artifact of a single outlier, RoDyn uniformly secures the SOTA position across all six individual sub-tasks delivering maximum absolute performance gain of approximately 40\% over other methods on hammer task. 

\subsection{Ablation Study}
\label{sec:ab}
To systematically validate our architectural design choices, we conduct rigorous ablation studies on the BAIR dataset~\cite{ebert2017self}. All model variants are trained under strictly identical configurations, and inference times are benchmarked on a single NVIDIA RTX 3090 GPU to ensure fair comparisons. The quantitative results are detailed in \Cref{tab:ab}.

\noindent \textbf{Tokenizer Efficiency.} 
We compare it against Naive Concat baseline, which simply concatenates the multi-modal observations along the sequence dimension, yielding a transformer input sequence three times longer. 
In sharp contrast, our deep cross-modal fusion gracefully compresses the 2.5D state space into a unified representation. 
This design accelerates inference by 86$\times$ (down to 10s) while stabilizing the temporal generation quality.

\noindent \textbf{Module Integration.} Next, we isolate the performance gains contributed by our core modules. 
Starting from a pure 2D visual Baseline, we incrementally integrate the Robot-Dynamic Tokenizer and the Mask-Guided AR Transformer. 
As shown in the lower section of \Cref{tab:ab}, these ablations show that both the geometry-aware latent space and the physically constrained causal generation contribute to RoDyn.

\begin{table}[tb]
\centering
\caption{\textbf{Ablation Study on the BAIR Dataset.} 
The top section shows the critical role of the RD Tokenizer in accelerating inference. The bottom section isolates the gains from integrating our physical reasoning modules.}
\resizebox{\linewidth}{!}{
\begin{tabular}{lccccc}
\toprule
\textbf{Variant} & FVD$\downarrow$ & PSNR$\uparrow$ & SSIM$\uparrow$ & LPIPS$\downarrow$ & Time \\
\midrule
\multicolumn{6}{c}{\textit{Tokenizer Efficiency}} \\
\midrule
Naive Concat & 739.2 & 16.64 & 0.735 & 0.176 & 860s \\
\textbf{RD Tokenizer} & \textbf{67.6} & \textbf{23.22} & \textbf{0.885} & \textbf{0.056} & \textbf{10s} \\
\midrule
\multicolumn{6}{c}{\textit{Module Integration}} \\
\midrule
Baseline (2D)    & 70.2 & 22.50 & 0.865 & 0.066 & - \\
+ RD Tokenizer   & 67.5 & 22.67 & 0.873 & 0.061 & - \\
+ Mask-Guided AR & 70.6 & 22.84 & 0.877 & 0.059 & - \\
\textbf{RoDyn (Full)} & 67.6 & \textbf{23.22} & \textbf{0.885} & \textbf{0.056} & - \\
\bottomrule
\end{tabular}
}
\label{tab:ab}
\vspace{-0.3cm}
\end{table}

\section{Conclusion}
In this work, we introduce RoDyn, a novel Robot-Dynamic 2.5D World Model tailored for complex robotic manipulation. 
Rather than relying on physically agnostic 2D pixels, our method explicitly incorporates spatial topologies and kinematic priors to construct a geometry-aware and physically grounded state space. To achieve highly efficient representation, we propose the Robot-Dynamic Tokenizer, which aligns multi-modal observations into unified, physics-aware tokens. 
The temporal dynamics are then sequentially modeled via a Mask-guided Autoregressive Transformer that strictly anchors future state generation on causal robotic actions. 
Extensive experiments demonstrate that RoDyn not only establishes new sota visual generation fidelity across large-scale datasets, but also translates these geometric capabilities into substantial performance leaps for downstream robotic tasks, significantly enhancing visual model-based reinforcement learning and real-world imitation learning. Nevertheless, scaling up the pre-training data across more diverse robotic embodiments and heterogeneous action spaces remains a promising direction to fully unlock the universal generalizability of 2.5D multi-modal world models.






\section*{ACKNOWLEDGMENT}

This work was supported by the Singapore National Robotics Programme under Research Projects DS-RFM M25N4N2009 and DEM M25N4N2150.


\bibliographystyle{IEEEtran}
\bibliography{ref}

\end{document}